\documentclass{article}

  \PassOptionsToPackage{numbers, compress}{natbib}
\usepackage{etoolbox}
\usepackage[belowskip=-9pt,aboveskip=0pt]{caption}
\usepackage[final]{nips_2017}

\usepackage[utf8]{inputenc} 
\usepackage[T1]{fontenc}    
\usepackage{hyperref}       
\usepackage{url}            
\usepackage{booktabs}       
\usepackage{amsfonts}       
\usepackage{nicefrac}       
\usepackage{microtype}      
\usepackage{epsfig}
\usepackage{graphicx}
\usepackage{subcaption}
\usepackage{multirow}
\usepackage{amsmath}
\usepackage{amssymb}
\usepackage{wrapfig}
\graphicspath{{figures/}{../figures/}}

\hypersetup{
    colorlinks=true,
    linkcolor=blue,
    filecolor=magenta,      
    urlcolor=cyan,
}
 
\title{Learning Pixel-Distribution Prior with Wider Convolution for Image Denoising}

\author{
   Peng Liu
   \\
  University of Florida\\
  \texttt{pliu1@ufl.edu} \\
    \And
    Ruogu Fang\\
    University of Florida \\
    \texttt{ruogu.fang@bme.ufl.edu} \\
}

\begin{document}

\maketitle

\begin{abstract}

In this work, we explore an innovative strategy for image denoising 
by using convolutional neural networks (CNN) to learn pixel-distribution from noisy data. 
By increasing CNN's width with large reception fields and more channels in each layer, 
CNNs can reveal the ability of learning pixel-distribution, which is a prior excising in many different types of noise. 
The key to our approach is a discovery that wider CNNs tends to learn the pixel-distribution features, which provides the probability of that inference-mapping primarily relies on the priors instead of deeper CNNs with more stacked non-linear layers. 
We evaluate our work: Wide inference Networks (WIN) on additive white Gaussian noise (AWGN) and demonstrate that by learning the pixel-distribution in images, WIN-based network consistently achieves significantly better performance than current state-of-the-art deep CNN-based methods in both quantitative and visual evaluations. \textit{Code and models are available at \url{https://github.com/cswin/WIN}}.

\end{abstract}

\section{Prior: pixel-distribution features} 

\begin{figure}[ht]
\centering
  \includegraphics[scale=0.4]{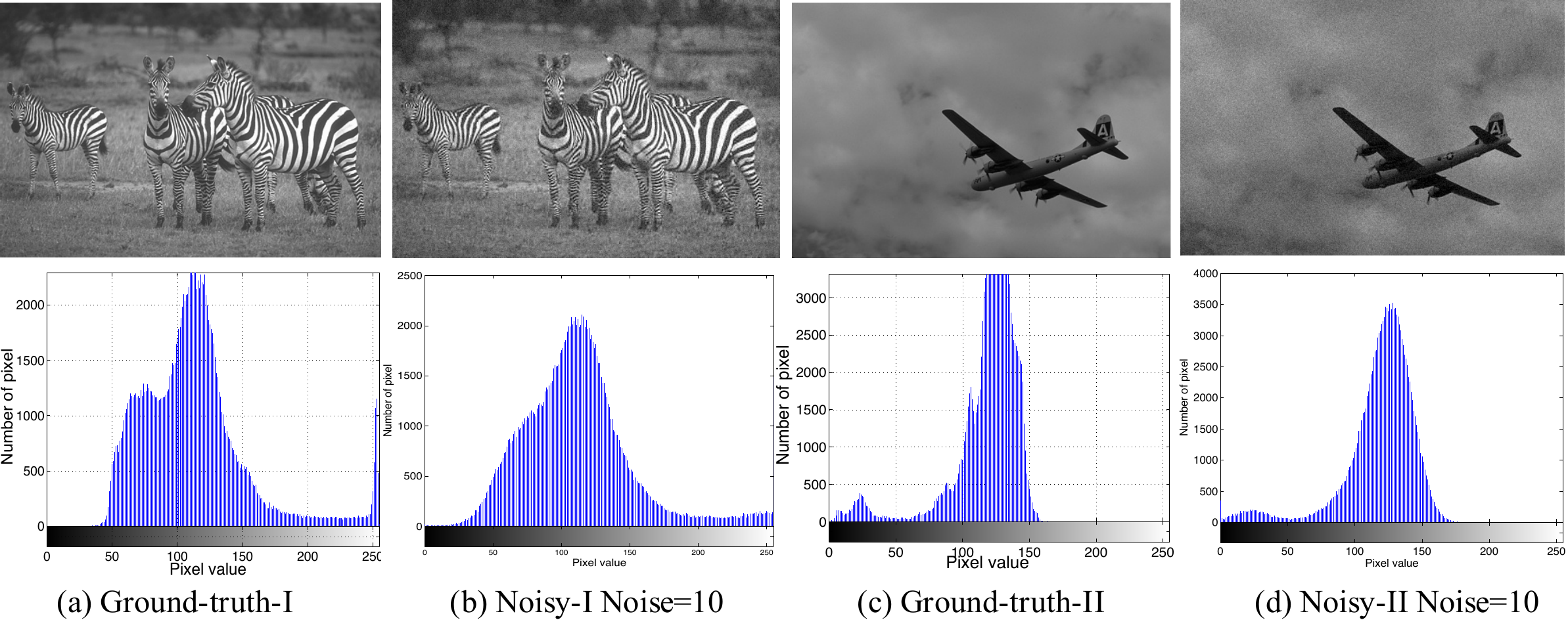}
  \caption{\small Compare the pixel distributions of histograms of two different
  images added additive white Gaussian noise (AWGN) with same noise level $\sigma=10$. }
\label{fig:histogram_N10}
\end{figure}

\begin{figure}[ht]
\centering
  \includegraphics[scale=0.4]{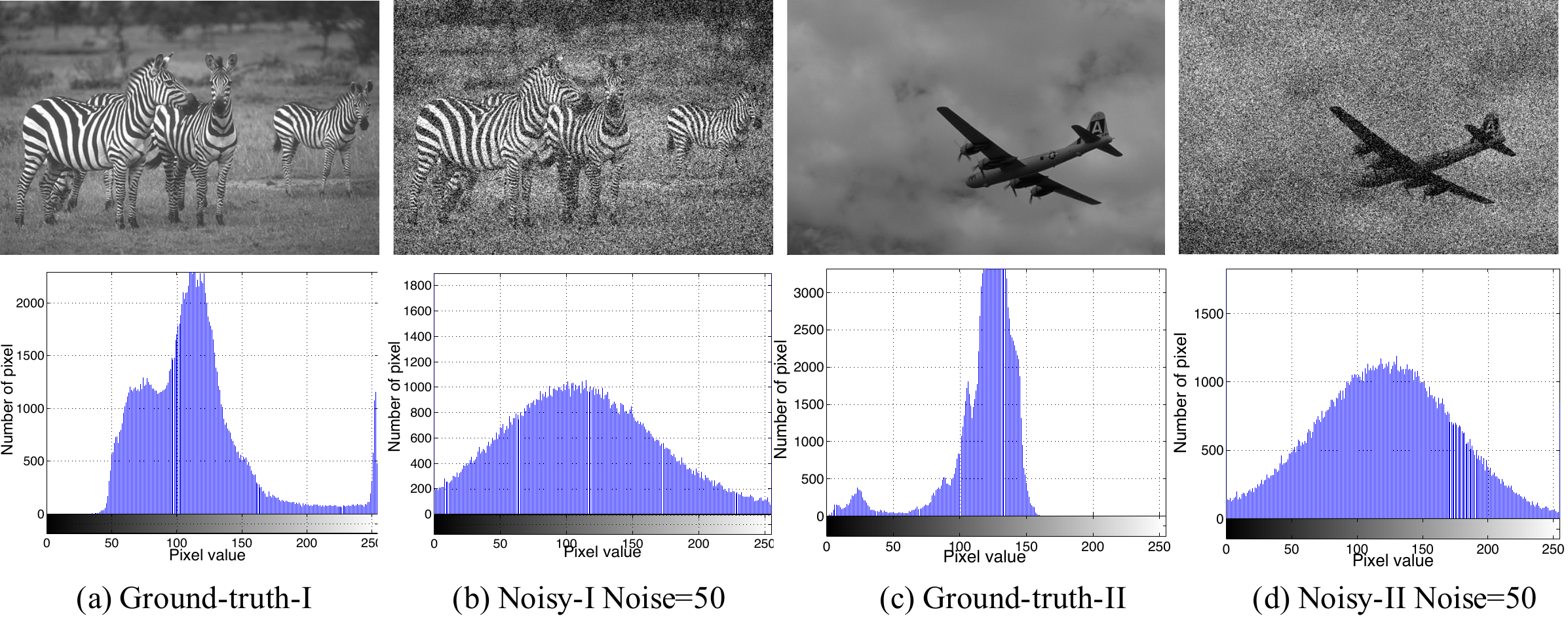}
  \caption{\small Compare the pixel distributions of histograms of two different
  images added additive white Gaussian noise (AWGN) with same noise level $\sigma=50$. }
  \vspace{-1em}
\label{fig:histogram_N50}
\end{figure}

In low-level vision problems, pixel-level features are the most important features. 
We compare the histograms of different images in various noise levels to investigate the pixel-level
features having a certain of consistency. As we can see from Fig.~\ref{fig:histogram_N10} and Fig.~\ref{fig:histogram_N50}, 
the pixel-distribution in noisy images is more similar in higher noise level $\sigma=50$ than lower noise level $\sigma=10$. 

WIN inferences noise-free images based on the learned pixel-distribution features. When the noise level is the higher, the pixel-distribution features are more similar. Thus, WIN can learn more pixel-distribution features from noisy images having higher level noise. This is the reason that WIN performs even better in higher-level noise, which can be seen and verified in section \ref{morevisual}.

\section{Wider Convolution Inference Strategy}

\begin{wrapfigure}{r}{0.4\textwidth}
  \centering 
    \includegraphics[width=0.38\textwidth]{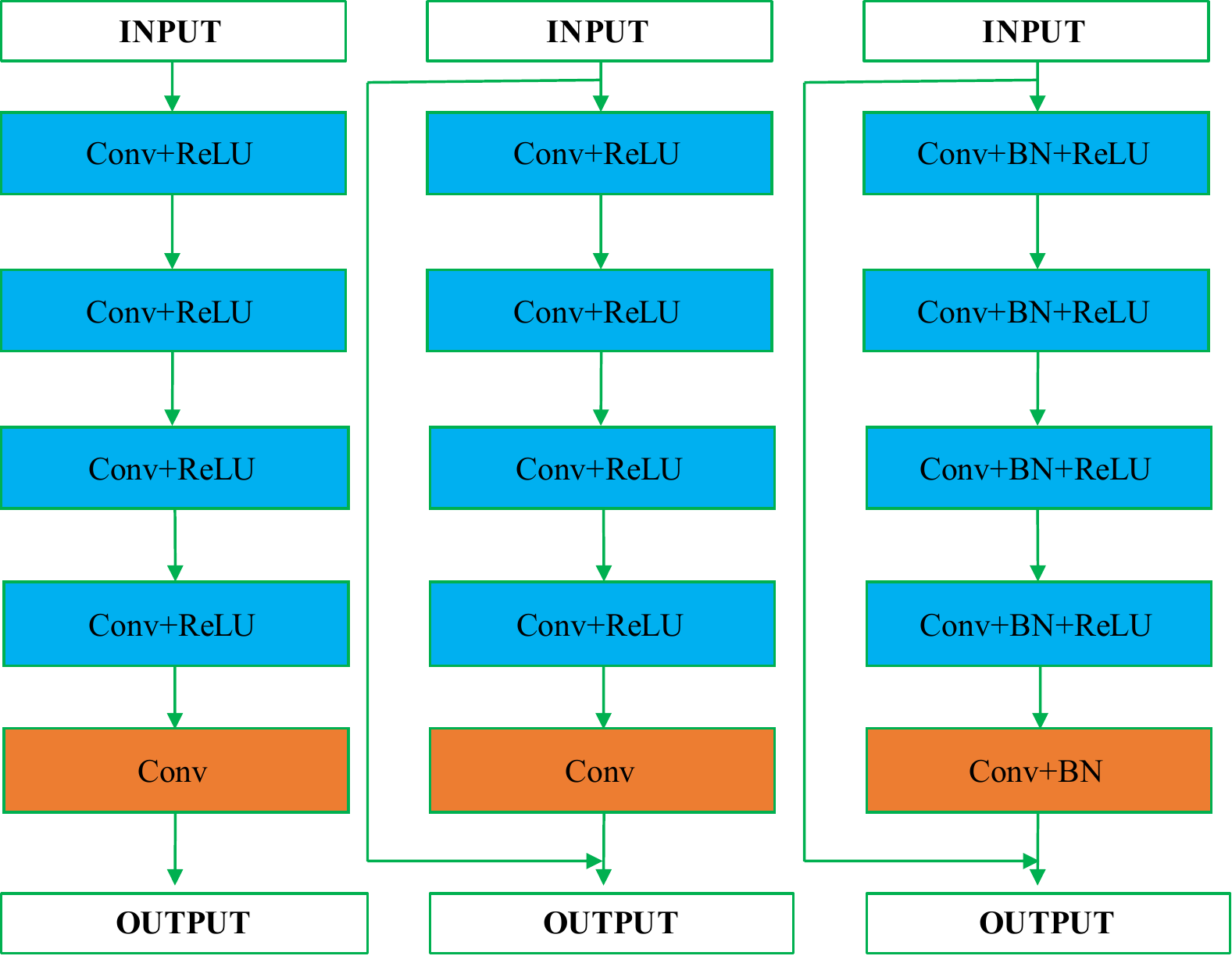}
  \caption{ Architectures (a) WIN5  (b) WIN5-R (c) WIN5-RB.}
  \label{fig:architecture}
\end{wrapfigure}

In Fig.~\ref{fig:architecture}, we illustrate the architectures of WIN5, WIN5-R, and WIN5-RB.

\subsection{Architectures} 

Three proposed models have the identical basic structure: $L=5$ layers and $K=128$ filters of size $F=7\times7$ in most convolution layers, except for the last one with $K=1$ filter of size $F=7\times7$. The differences among them are whether batch normalization (BN) and an input-to-output skip connection are involved. WIN5-RB has two types of layers with two different colors. (a) Conv+BN+ReLU \cite{nair2010rectified}: for layers 1 to $L-1$, BN is added between Conv and ReLU \cite{nair2010rectified}. (b) Conv+BN: for the last layer,
$K=1$ filters of size $F=7\times7$ is used to reconstruct the \(R(y)\approx\displaystyle -n\). In addition, a shortcut skip connecting the input (data layer) with the output (last layer) is added to merge the input data with \(R(y)\) as the final recovered image. 

\subsection{Having Knowledge Base with Batch-Normal}
In this work, \textit{we employ Batch Normalization (BN) for extracting pixel-distribution
statistic features and reserving training data means and variances in networks for denoising 
inference instead of 
using the regularizing effect of improving the generalization of a learned model.}

The regularizer-BN can keep the data distribution the same as input: Gaussian distribution. 
This distribution consistency between input and regularizer ensures more pixel-distribution
statistic features can be extracted accurately. 
The integration of BN \cite{ioffe2015batch} into more filters will
further preserve the prior information of the training set. Actually, a number of state-of-the-art studies \cite{elad2006image, joshi2009image,xu2015patch} have adopted image priors (e.g. distribution statistic information) to achieve impressive performance.

\textbf{Can Batch Normalization work without a Skip Connection?} 
In WINs, BN \cite{ioffe2015batch} cannot work without the input-to-output skip connection and is always over-fitting. 
In WIN5-RB's training, BN keeps the distribution of input data consistent and the skip connection
can not only introduce residual learning but also guide the network to extract the certain features in common: pixel-distribution. 
Without the input data as a comparison, BN could bring negative effects by keeping the each input distribution same, especially, when a task is to output pixel-level feature map. In DnCNN, two BN layers are removed from the first and last layers,
by which a certain degree of the BN's negative effects can be reduced.
Meantime DnCNN also highlights network's generalization ability largely
relies on the depth of networks. 

\begin{figure}[ht]
\centering
  \includegraphics[scale=0.5]{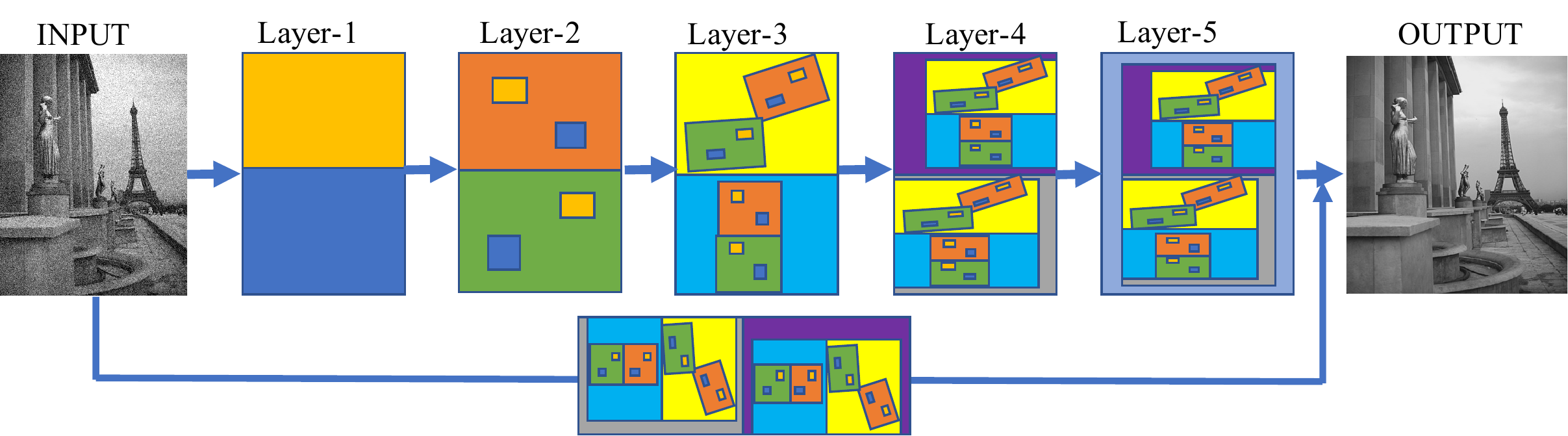}
  \caption{The process of denoising inference by sparse distribution statistics features. Learned priors (means and variances) are preserved in WINs as knowledge base for denoising inference. When WIN has more channels to preserve more data means and variances, various combinations of these feature maps can corporate with residual learning to infer the noise-free images more accurately. }
\label{fig:workflow}
\end{figure}

In Fig.\ref{fig:workflow}, learned priors (means and variances) are preserved in WINs as knowledge base for denoising inference. When WIN has more channels to preserve more data means and variances, various combinations of these feature maps can corporate with residual learning to infer the noise-free images more accurately.

\section{Experimental Results} \label{morevisual}
  \begin{table}[ht]
\scriptsize
\caption{\small The average results of PSNR (dB) / SSIM / Run Time (seconds) of different methods on the BSD200-test~\cite{MartinFTM01} (200 images). Note: WIN5-RB-B (blind denoising)  is trained on larger number of patches as data augmentation is adapted.This is the reason why WIN5-RB-B (trained on $\sigma=[0-70]$) can outperform WIN5-RB-S (trained on single $\sigma=10, 30, 50, 70$ separately) in some cases. }
\label{tab:BSD-test}
\centering
  \begin{tabular}{llllllll}
    \toprule
   \multicolumn{8}{c}{PSNR (dB) / SSIM }                   \\
    \cmidrule{1-8}
    $\sigma$ & BM3D \cite{dabov2009bm3d}   & RED-Net \cite{mao2016image} & DnCNN \cite{zhang2016beyond} & WIN5 & WIN5-R & WIN5-RB-S & WIN5-RB-B\\
     
    \midrule
    10  & 34.02/0.9182 & 32.96/0.8963 & 34.60/0.9283  & 34.10/0.9205  & 34.43/0.9243  &  \textbf{35.83/0.9494}  & 35.43/0.9461 \\
    30  & 28.57/0.7823 & 29.05/0.8049 & 29.13/0.8060  &28.93/0.7987 & 30.94/0.8644 & \textbf{33.62/0.9193}  & 33.27/0.9263    \\
    50  & 26.44/0.7028 & 26.88/0.7230 & 26.99/0.7289          &28.57/0.7979 & 29.38/0.8251 &31.79/0.8831 & \textbf{32.18/0.9136}\\
    70  & 25.23/0.6522 & 26.66/0.7108 & 25.65/0.6709          &27.98/0.7875 & 28.16/0.7784 & 30.34/0.8362  & \textbf{31.07/0.8962}\\
    \bottomrule
      
   \multicolumn{8}{c}{Run Time(s)}                   \\
 
   \midrule
   
    30  &\textbf{1.67} &69.25  & 13.61 & 15.36 & 15.78 &  20.39  &  15.82 \\
    50  &\textbf{2.87} &70.34  & 13.76 & 16.70 & 22.72 &  21.79  &  13.79    \\
    70  &\textbf{2.93} &69.99  & 12.88 & 16.10 & 19.28 &  20.86  &  13.17   \\
    \bottomrule
  \end{tabular}
   \vspace{-2em}
\end{table}

\begin{wrapfigure}{r}{0.3\textwidth}
\vspace{-1em}
 \footnotesize
\centering
  \includegraphics[width=0.28\textwidth]{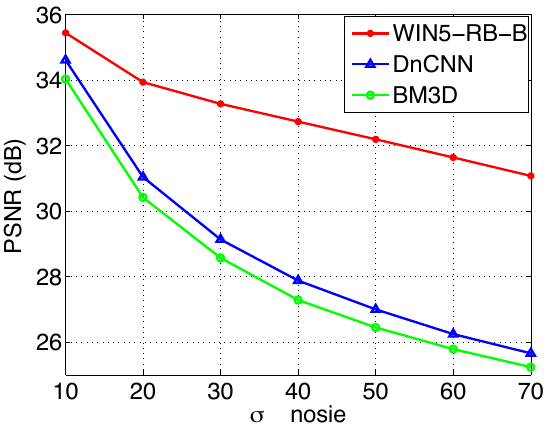}
  \caption{\small 
  Behavior at different noise levels of average PSNR on BSD200-test. WIN5-RB-B (blind denoising) is 
  trained for $\sigma=[0-70]$ and outperforms BM3D~\cite{dabov2009bm3d} and DnCNN~\cite{zhang2016beyond} on all noise levels and is significantly more stable even on higher noise levels.
  }
 \vspace{-1em}
\label{fig:behavior-blind}
\end{wrapfigure}
\subsection{Quantitative Result}
 The quantitative result on test set BSD200 is shown on Table~\ref{tab:BSD-test} including noise levels $\sigma=10, 30, 50, 70$.
 Moreover, we compare WIN5-RB-B, DnCNN and BM3D behaviors at different noise levels of average PSNR on BSD200-test. As we can see from Fig.\ref{fig:behavior-blind}, WIN5-RB-B (blind denoising) trained for $\sigma=[0-70]$ outperforms BM3D~\cite{dabov2009bm3d} and DnCNN~\cite{zhang2016beyond} on all noise levels and is significantly more stable even on higher noise levels. 
 
 In addition, in Fig.\ref{fig:behavior-blind}, as the noise level is increasing, the performance gain of WIN5-RB-B is getting larger, while the performance gain of DnCNN comparing to BM3D is not changing much as the noise level is changing. Compared with WINs, DnCNN is composed of even more layers embedded with BN. This observation indicates that the performance gain achieved by WIN5-RB does not mostly come from BN's regularization effect but the pixel-distribution features learned and relevant priors such as means and variances reserved in WINs. Both Larger kernels and more channels can promote CNNs more likely to learn pixel-distribution features.

\subsection{Visual results}
For Visual results, We have various images from two different datasets,
BSD200-test and Set12, with noise levels $\sigma=10, 30, 50, 70$ applied separately.

\textbf{One image from BSD200-test with noise level=10 }

\begin{figure}[ht]
\centering
  \includegraphics[width=\textwidth]{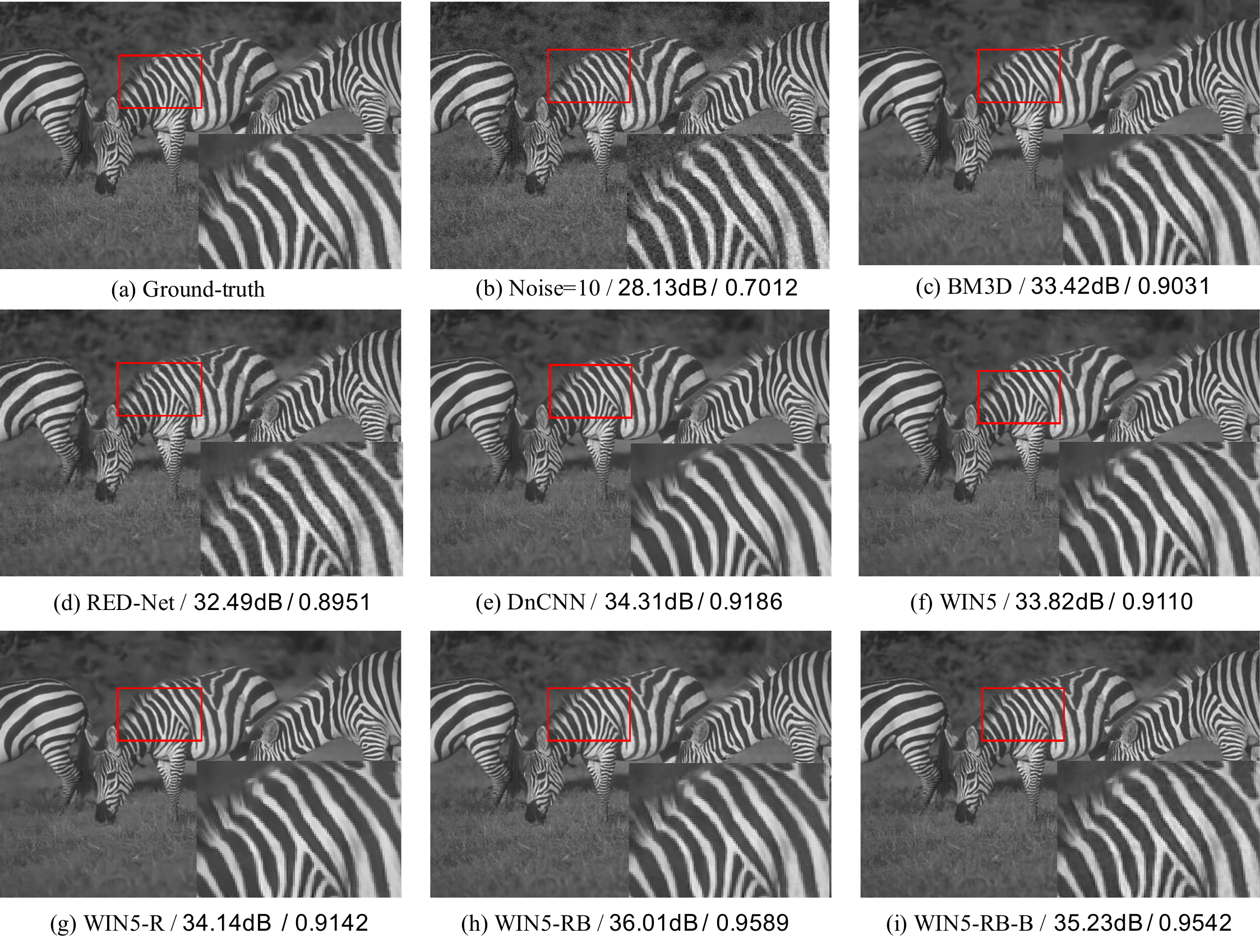}
  \caption{Visual results of one image from BSD200-test with noise level $\sigma=10$ 
  along with PSNR(dB) / SSIM. As we can see, our proposed methods 
  can yield more natural and accurate details 
  in the texture as well as visually pleasant results.}
\label{fig:visual00}
\end{figure}
\newpage


\textbf{One image from BSD200-test with noise level=30}
\begin{figure}[ht]
\centering
  \includegraphics[width=\textwidth]{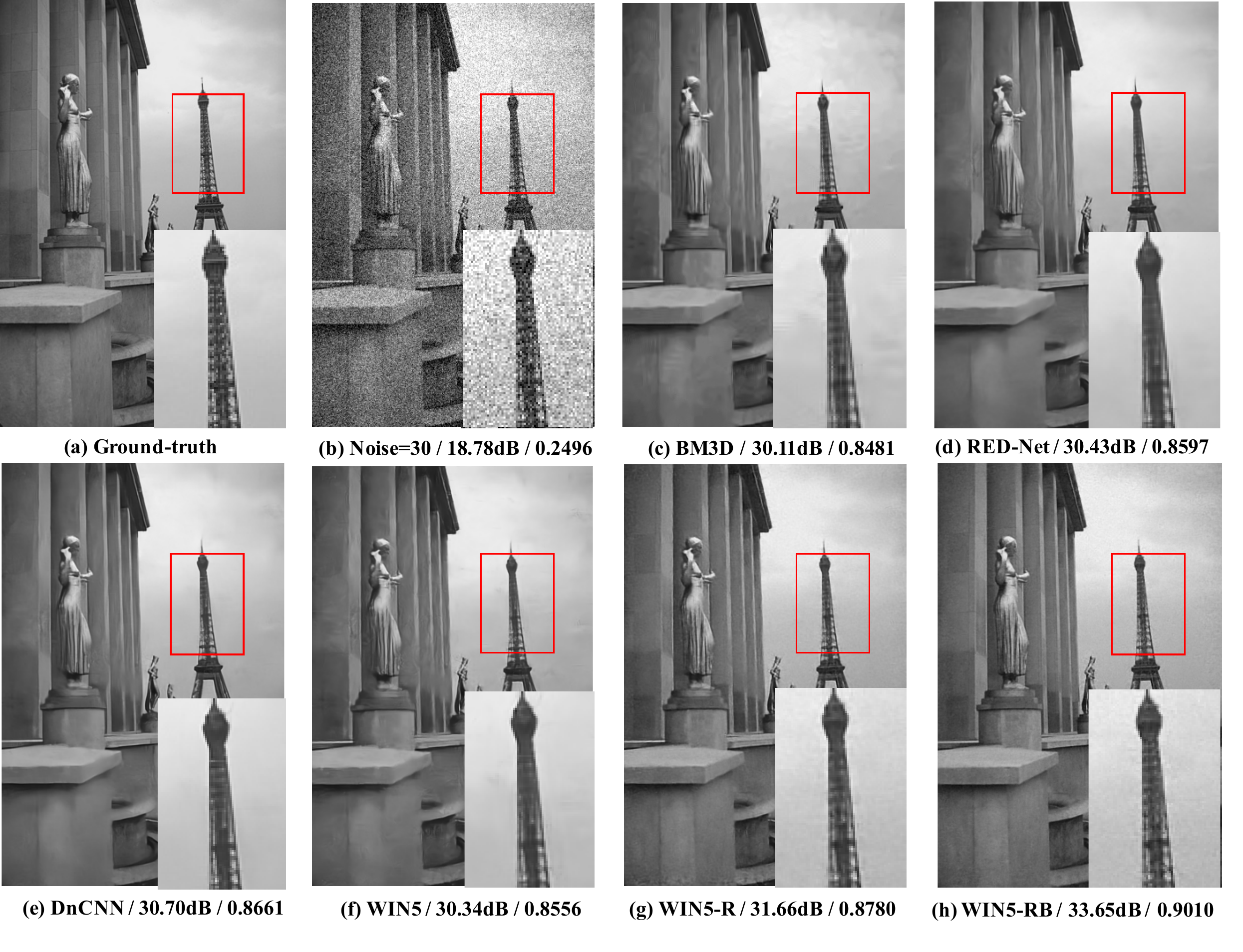}
  \caption{Visual results of one image from BSD200-test with noise level $\sigma=30$ 
  along with PSNR(dB) / SSIM. As we can see, our proposed methods 
  can yield more natural and accurate details 
  in the texture as well as visually pleasant results.}
\label{fig:visual5}
\end{figure}
\newpage
\textbf{One image from BSD200-test with noise level=50}
\begin{figure}[ht]
\centering
  \includegraphics[width=\textwidth]{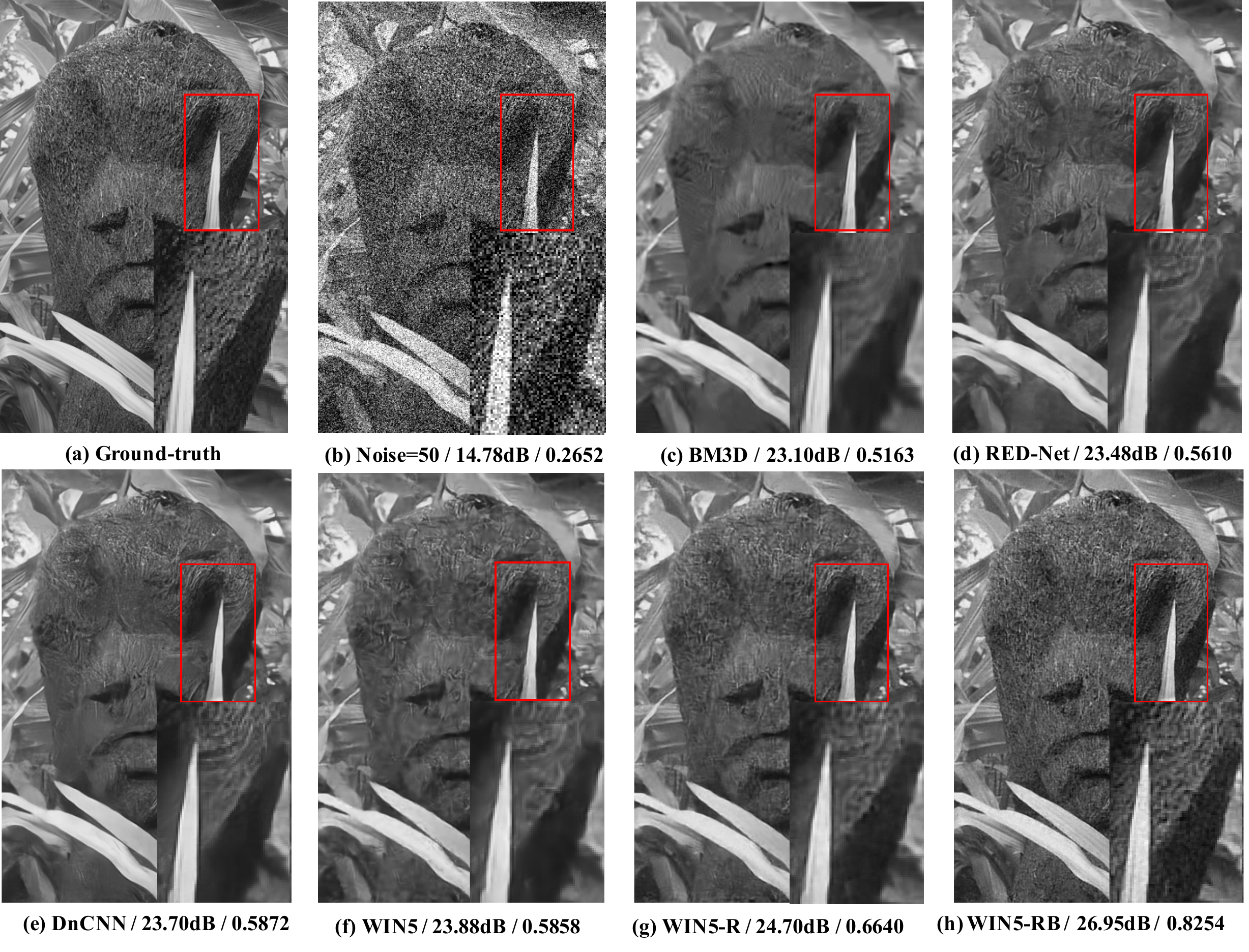}
  \caption{Visual results of one image from BSD200-test with noise level $\sigma=50$ 
  along with PSNR(dB) / SSIM. As we can see, our proposed methods 
  can yield more natural and accurate details 
  in the texture as well as visually pleasant results.}
\label{fig:visual2}
\end{figure}

\newpage

\textbf{One image from BSD200-test with noise level=70 }
\begin{figure}[ht]
\centering
  \includegraphics[width=\textwidth]{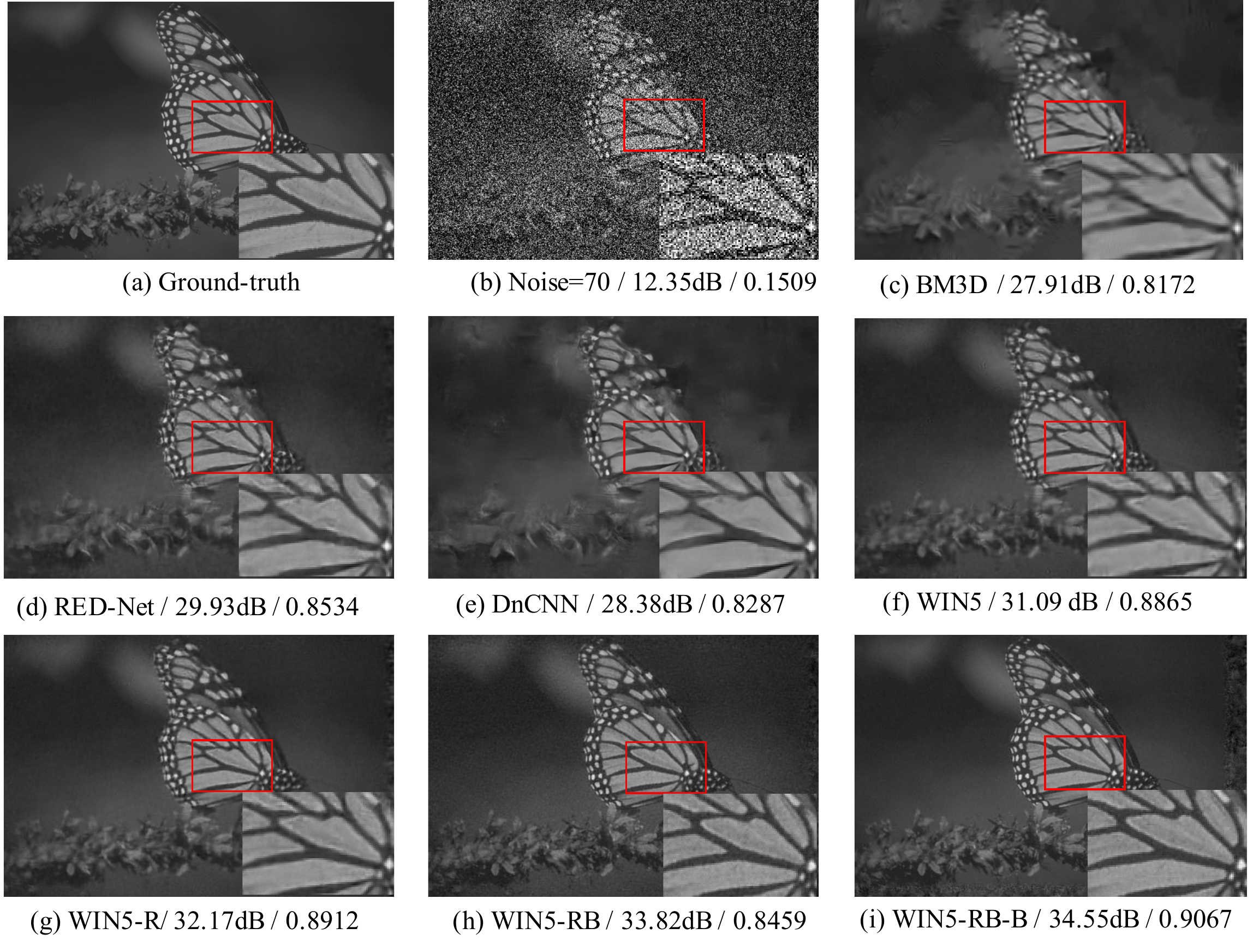}
  \caption{Visual results of one image from BSD200-test with noise level $\sigma=70$ 
  along with PSNR(dB) / SSIM. As we can see, our proposed methods 
  can yield more natural and accurate details 
  in the texture as well as visually pleasant results.}
\label{fig:visual1}
\end{figure}

\newpage

\textbf{One image from Set12 with noise level=10}
\begin{figure}[ht]
\centering
  \includegraphics[width=\textwidth]{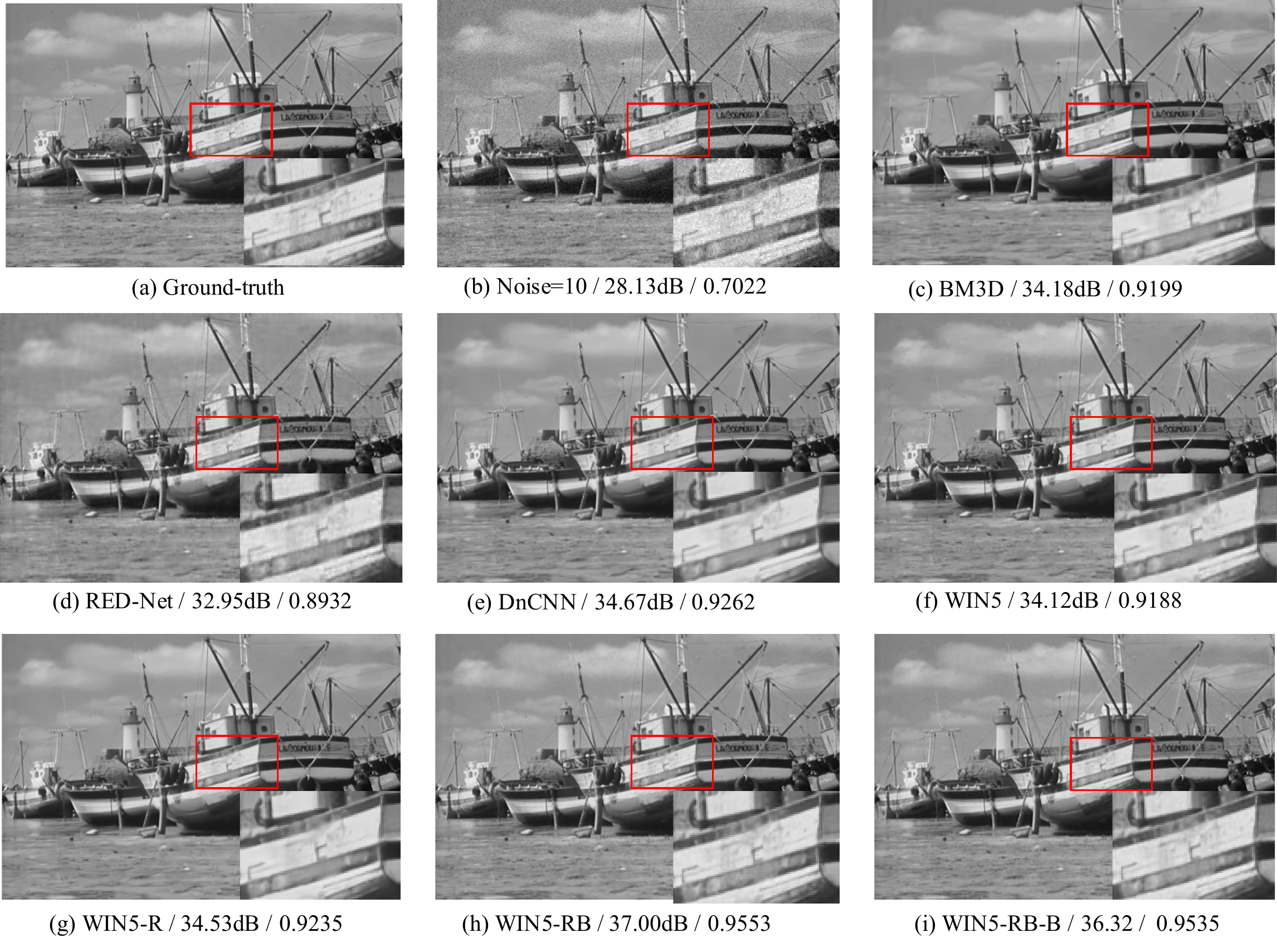}
  \caption{Visual results of one image from Set12 with noise level $\sigma=10$ 
  along with PSNR(dB) / SSIM. As we can see, our proposed methods 
  can yield more natural and accurate details 
  in the texture as well as visually pleasant results.}
\label{fig:visual7}
\end{figure}

\newpage
\textbf{One image from Set12 with noise level=30}

\begin{figure}[ht]
\centering
  \includegraphics[width=\textwidth]{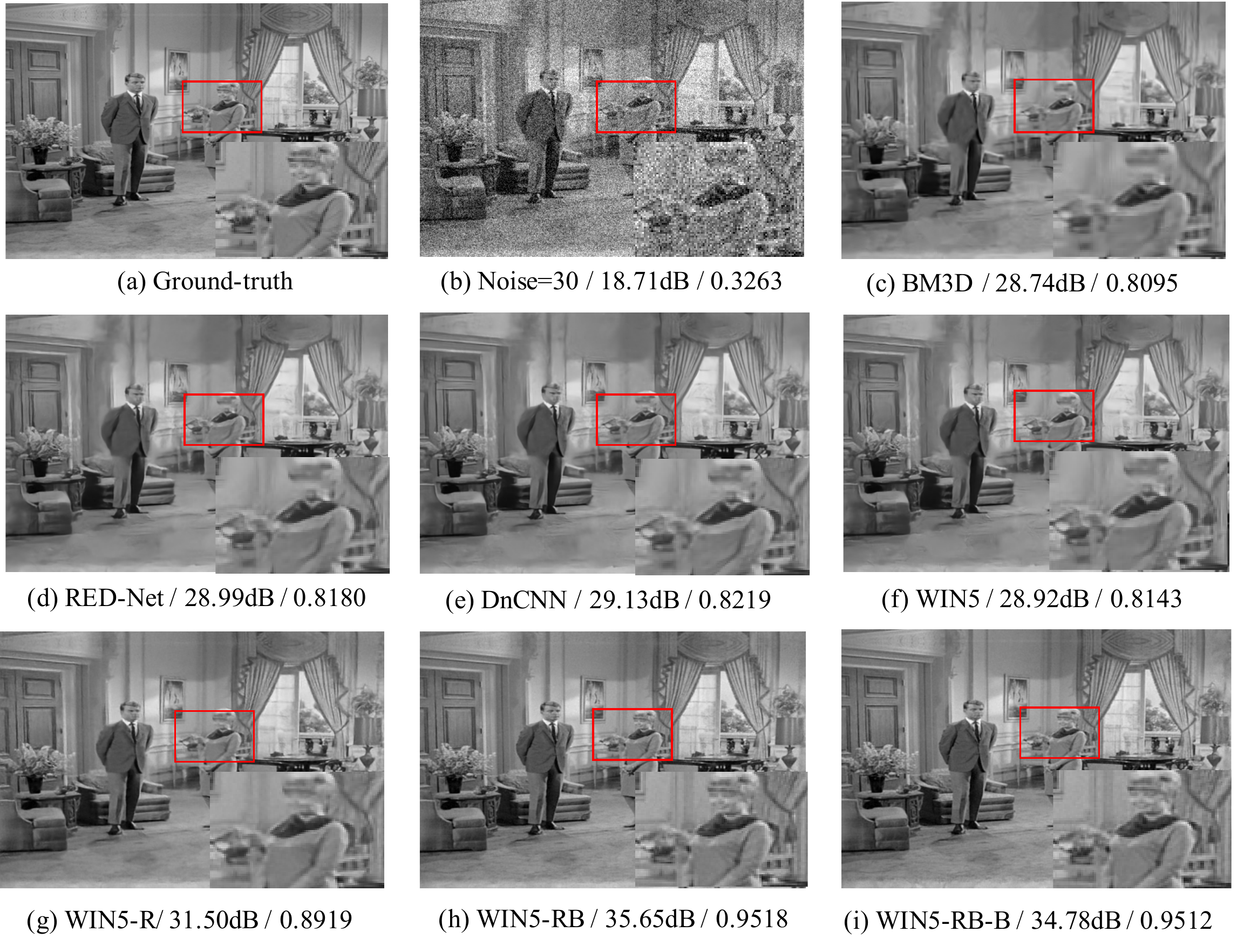}
  \caption{Visual results of one image from Set12 with noise level $\sigma=30$ 
  along with PSNR(dB) / SSIM. As we can see, our proposed methods 
  can yield more natural and accurate details 
  in the texture as well as visually pleasant results.}
\label{fig:visual6}
\end{figure}
\newpage

\textbf{One image from Set12 with noise level=50}

\begin{figure}[ht]
\centering
  \includegraphics[width=\textwidth]{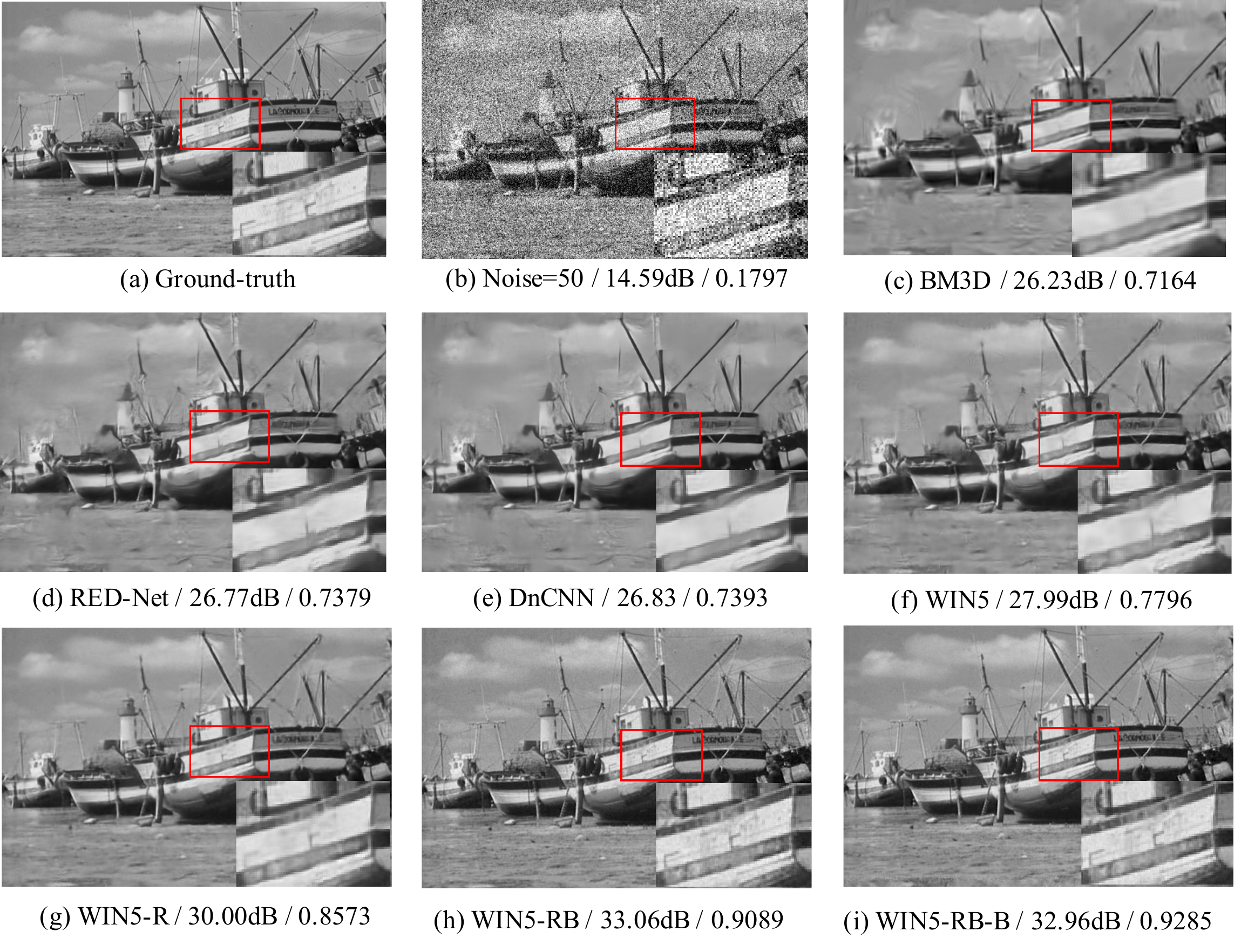}
  \caption{Visual results of one image from Set12 with noise level $\sigma=50$ 
  along with PSNR(dB) / SSIM. As we can see, our proposed methods 
  can yield more natural and accurate details 
  in the texture as well as visually pleasant results.}
\label{fig:visual3}
\end{figure}

\newpage

{\small
\bibliographystyle{ieee}
\bibliography{refer_WIN}
}

\end{document}